\documentclass[11pt,a4paper]{article}
\usepackage[T1]{fontenc}
\usepackage[utf8]{inputenc}
\usepackage{lmodern,microtype}
\usepackage[margin=24mm]{geometry}
\usepackage{amsmath,amssymb,booktabs,array,graphicx}
\usepackage[font=small,labelfont=bf]{caption}
\usepackage{placeins}
\usepackage[colorlinks=true,linkcolor=black,citecolor=black,urlcolor=blue,pdfauthor={Bin Yin},pdftitle={Online State-Representation Adaptation under Synthetic Interaction-Cue Remapping}]{hyperref}
\newcommand{\AdaptiveLate}{\ensuremath{93.39}}
\newcommand{\FrozenLate}{\ensuremath{64.63}}
\newcommand{\DirectLate}{\ensuremath{95.65}}
\newcommand{\BanditLate}{\ensuremath{94.23}}
\newcommand{\AdaptiveUtility}{\ensuremath{94.10}}
\newcommand{\BanditUtility}{\ensuremath{98.03}}
\newcommand{\PlasticLate}{\ensuremath{28.76\,[8.73, 53.19]}}
\newcommand{\PlasticUtility}{\ensuremath{80.71\,[13.23, 166.42]}}
\newcommand{\DirectLateDiff}{\ensuremath{-2.26\,[-2.76, -1.71]}}
\newcommand{\DirectUtilityDiff}{\ensuremath{0.92\,[-1.60, 4.15]}}
\newcommand{\BanditLateDiff}{\ensuremath{-0.84\,[-1.45, -0.37]}}
\newcommand{\BanditUtilityDiff}{\ensuremath{-3.93\,[-7.17, -0.89]}}
\newcommand{\HistoryLateDiff}{\ensuremath{-1.15\,[-1.52, -0.79]}}
\newcommand{\HistoryUtilityDiff}{\ensuremath{-3.45\,[-4.48, -2.43]}}
\newcommand{\FixedUtilityDiff}{\ensuremath{-2.55\,[-2.93, -2.09]}}
\newcommand{\MatchedUtilityDiff}{\ensuremath{2.22\,[0.20, 4.81]}}
\newcommand{\AdaptiveUnsafe}{\ensuremath{3.09}}
\newcommand{\DirectUnsafe}{\ensuremath{4.12}}
\newcommand{\AdaptiveMissed}{\ensuremath{10.01}}
\newcommand{\DirectMissed}{\ensuremath{4.62}}
\newcommand{\ValueLate}{\ensuremath{87.87}}
\newcommand{\ValueMissed}{\ensuremath{20.95}}
\newcommand{\AdaptiveBrier}{\ensuremath{0.0500}}
\newcommand{\FrozenBrier}{\ensuremath{0.2473}}

\title{\vspace{-1.2cm}Online State-Representation Adaptation under Synthetic Interaction-Cue Remapping}
\author{Bin Yin\\[2pt]\small School of Psychology, Fujian Normal University}
\date{}
\begin{document}
\maketitle
\begin{abstract}
When cues change what they predict, continuing to update a learned mapping may help an agent choose the appropriate action. We tested this in a one-step synthetic task by comparing a logistic state predictor that continued to learn from state labels with the same predictor frozen at transfer. We evaluated ten conditions in six held-out environments, with 1,296 paired runs per condition and 1,296,000 transfer records in total. Late action accuracy was \AdaptiveLate\% with online updates and \FrozenLate\% with frozen coefficients, a paired difference of \PlasticLate\ percentage points (95\% environment-bootstrap interval). A direct policy reached \DirectLate\% using the same predictions with a different action threshold. A contextual bandit reached \BanditLate\% and earned more cumulative utility than the adaptive predictor. The bandit used the reward of its chosen action, which in this task reveals the true state when combined with that action. Weights based on development history lowered cumulative utility relative to neutral weights (difference \HistoryUtilityDiff). The adaptive predictor outperformed its frozen counterpart in late accuracy in each of the six test environments. The other comparisons show how the choice of action threshold and the information available in feedback affect what can be attributed to online adaptation.
\end{abstract}

\section{Introduction}
A cue that predicts support during training may predict threat after deployment. This change in the association between observations and the target state is a form of concept drift \cite{gama}. An online learner can respond by revising its state predictions as new labels arrive. In a decision task, however, action performance also depends on how an agent translates its predictions into actions. Measuring both state prediction and action performance helps identify which part of the learner changed.

We test this distinction in a synthetic task where an agent chooses whether to approach or avoid after observing six cues. At transfer, the cues can change their association with the underlying state while the action rule and exploration schedule remain the same for the adaptive and frozen predictors. The main comparison asks how much continuing the same supervised update improves performance. A second comparison uses the same predictions across learners and changes the decision threshold. Two value learners then show how performance varies when algorithms learn from different feedback.

The task also separates early exposure from the costs of current actions. Development histories differ in how often the supportive state occurs. Each history is crossed with each current payoff profile, so a learner's decision rule can be based on the current payoff, the earlier state frequencies, or a fixed reference. This design makes it possible to ask whether the tested history rule improves decisions on the same transfer streams.

\section{Methods}
\subsection{Task and transfer design}
In each episode, the simulator samples a binary state $s$, where $s=0$ denotes threat and $s=1$ denotes support. It then generates a vector of six Gaussian cues $x$. The learner observes $x$ and chooses either avoidance ($a=0$) or approach ($a=1$). The simulator scores the decision using the true state. An action changes that episode's payoff. Later states and cues are drawn independently of the learner's actions.

Each run starts with 160 development episodes and continues for 100 transfer episodes. During development, the probability of support is 0.8, 0.5, or 0.2, corresponding to supportive, ordinary, and adverse exposure histories. At transfer, the cue loadings and observation noise change. Depending on the environment, some cue loadings reverse sign or fall to zero, a previously weak cue becomes informative, and the support probability can drift. Environments also specify the probability of a feedback label flip and the feedback delay. Appendix~\ref{app:environment} gives the generator and the six test configurations.

The payoff weights $w_s$ and $w_a$ control the costs and benefits of actions in threat and support, respectively. Table~\ref{tab:payoff} gives the payoff $U(s,a)$ for a realized state and action. The three transfer profiles vary these weights independently of exposure history. Safety priority uses $(w_s,w_a)=(2.6,1.2)$, balanced uses $(2.0,1.5)$, and affiliation priority uses $(1.4,1.8)$. Each run uses one profile throughout transfer. Evaluation uses that profile even when a learner chooses actions using other weights.

\begin{table}[ht]
\centering\small
\caption{Payoff for each true state and action. The action that matches the state earns a positive payoff.}
\label{tab:payoff}
\begin{tabular}{@{}lcc@{}}
\toprule
State & Avoid ($a=0$) & Approach ($a=1$) \\
\midrule
Threat ($s=0$) & $0.45w_s$ & $-2w_s$ \\
Support ($s=1$) & $-0.35w_a$ & $w_a$ \\
\bottomrule
\end{tabular}
\end{table}

\subsection{Paired evaluation}
We crossed six test environments (IDs 200--205), three development histories, and three current payoff profiles, using 24 seed indices per environment and history. For each environment, history, and seed index, one random seed generates the states, cues, and label flips. A separate seed generates exploration and tie-breaking draws. The same streams are reused across the three payoff profiles and all model conditions.

The 432 distinct seed pairs therefore give $432\times3=1{,}296$ units for pairing model outcomes. One such unit is an environment, history, payoff profile, and seed index. Across ten model conditions and 100 transfer episodes per run, the archive contains 1,296,000 transfer records. The six environment configurations are the highest level of aggregation for the reported uncertainty.

\subsection{Learners and decision rules}
The adaptive state learner estimates the probability of support as $q(x)=\sigma(\beta^\top\phi)$, where $\phi=(1,x^\top)^\top$ and $\sigma$ is the logistic function. When an observed state label $\tilde{s}$ becomes available, the learner takes an update using the cue vector for that label and the current coefficients:
\begin{equation}
    \beta \leftarrow \beta + \eta \big[\tilde{s}-\sigma(\beta^\top\phi)\big]\phi .
\end{equation}
The frozen state learner has the same development training and keeps its development coefficients throughout transfer. Both learners choose the action with higher estimated expected utility under the current weights. For this rule, approach is preferred when $q(x)$ exceeds the threshold in Appendix~\ref{app:updates}.

The direct policy uses the same logistic update and the same label stream as the adaptive state learner. It uses $q(x)$ as a score for approach and applies a threshold of 0.5. The four weighting variants also share the adaptive state learner's predictions. They compute expected utility with equal weights, fixed balanced weights, weights derived from the realized fraction of true support states during development, or weights deliberately mismatched to the current profile. These variants change which action the learner takes at a given predicted probability.

The full-information value learner fits a linear value for each action. It calculates a target payoff for both actions from the observed state label $\tilde{s}$ and the phase's payoff weights. The contextual bandit instead uses the realized payoff of its chosen action to update that action's value. This is the chosen-action feedback interface of a contextual bandit \cite{li}. In this simulator, a positive payoff means that $a=s$ and a negative payoff means that $a\neq s$. Thus the sign and the chosen action identify the true state even when the label available to other learners has been flipped. The bandit update uses the reward directly.

At transfer, all nine learned conditions share exploration events with probability 0.05, the uniformly sampled actions on those events, and the draws used to break ties. Each action is chosen before using any feedback due in that episode. The tenth condition is a Bayes utility oracle. It uses the simulator's observation model and current support prior to estimate $q(x)$, then chooses the action with higher expected utility without exploration. The reported accuracy and utility of the oracle are realized outcomes on the same streams.

\subsection{Validation, outcomes, and uncertainty}
We used four separate validation environments (IDs 100--103) to select learning rates. Each candidate used eight seed indices per environment and history across all three payoff profiles, for 288 validation units. We evaluated three rates for the adaptive mapping and three for each value learner, selecting by mean cumulative utility with late accuracy as the first tie-breaker. Validation selected 0.12 for the adaptive mapping and 0.05 for each value learner. The frozen mapper, direct policy, and weighting variants used the adaptive mapper's rate.

An action is correct when $a=s$. Early accuracy uses transfer episodes 1--20 and late accuracy uses episodes 81--100. Cumulative utility sums realized payoffs over all 100 transfer episodes. For the late conditional error rates, we divide approaches in threat by the number of threat episodes and avoidances in support by the number of support episodes. We compute these ratios for each run before averaging. For methods that output $q(x)$, the Brier score measures mean squared probability error against true states over the full transfer period. Appendix~\ref{app:metrics} also defines the accuracy area and the first threshold crossing.

For a performance estimate, we average the run-level metric within each environment, then give the six environment means equal weight. For a contrast, we subtract the two models' scores within each paired unit before averaging the differences within environments. Each of 2,000 bootstrap replicates draws six environment means with replacement and averages them \cite{efron}. The interval gives the 2.5th and 97.5th percentiles of those averages. It describes variation when resampling the six fixed test configurations, with the pairing preserved within each configuration.

\clearpage
\section{Results}
\subsection{Online updates improve state prediction and late action accuracy}
The adaptive and frozen learners began transfer with the same cue-to-state mapping. Over episodes 81--100, the adaptive learner reached \AdaptiveLate\% accuracy and the frozen learner reached \FrozenLate\% (Table~\ref{tab:performance}). The paired difference was \PlasticLate\ percentage points and the cumulative utility difference was \PlasticUtility\ (Table~\ref{tab:contrasts}). The adaptive learner had higher late accuracy in all six environments, with gains that varied across configurations (Table~\ref{tab:environments}).

\begin{figure}[!ht]
\centering\includegraphics[width=\linewidth]{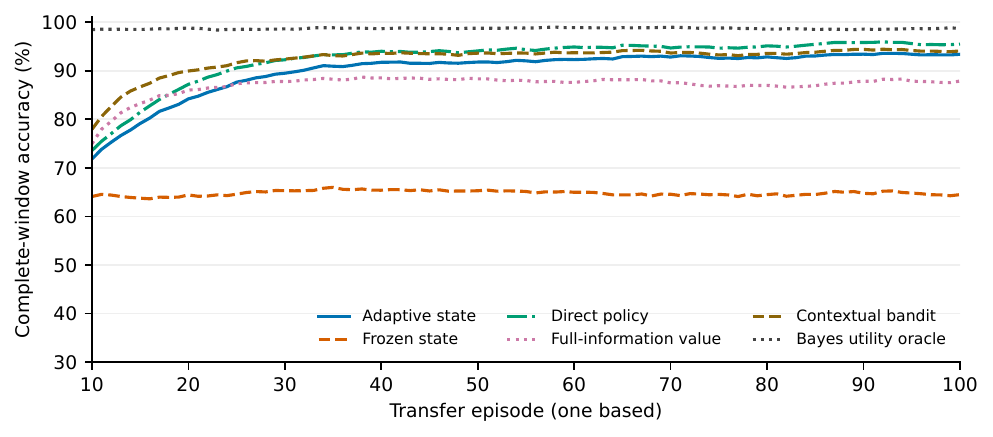}
\caption{Action accuracy during transfer. Each point summarizes the ten episodes ending at that position. Accuracy is averaged within each run, then within each environment and across the six environments. The four variants that change payoff weights are reported in the tables.}
\label{fig:curves}
\end{figure}

Mean Brier score was \AdaptiveBrier\ with online updates and \FrozenBrier\ with frozen coefficients (Table~\ref{tab:stateutility}). This improvement in state prediction accompanies the action gains under the shared payoff rule. The moving accuracy curves show when the average action performance separated after transfer (Figure~\ref{fig:curves}).

\begin{table}[!ht]
\centering\small
\caption{Late action accuracy and cumulative utility across the ten conditions. Entries are equally weighted environment means with 95\% environment-bootstrap intervals. Late accuracy uses episodes 81--100 and utility sums episodes 1--100.}
\label{tab:performance}
\begin{tabular}{@{}lrr@{}}
\toprule
Model condition & Late accuracy (\%) & Cumulative utility \\
\midrule
Adaptive state & $93.39\,[92.45, 94.36]$ & $94.10\,[81.90, 105.39]$ \\
Frozen state & $64.63\,[40.03, 85.07]$ & $13.39\,[-82.32, 91.26]$ \\
Direct policy & $95.65\,[94.69, 96.44]$ & $93.18\,[77.24, 106.82]$ \\
Full-information value & $87.87\,[86.15, 89.39]$ & $90.51\,[83.93, 96.89]$ \\
Contextual bandit & $94.23\,[93.17, 95.12]$ & $98.03\,[89.57, 106.11]$ \\
Neutral weights & $94.39\,[93.39, 95.22]$ & $94.57\,[80.77, 106.51]$ \\
Fixed balanced weights & $93.60\,[92.57, 94.58]$ & $93.67\,[82.07, 105.54]$ \\
History weights & $93.24\,[92.15, 94.38]$ & $91.12\,[78.65, 102.37]$ \\
Mismatched weights & $94.83\,[93.86, 95.60]$ & $91.88\,[77.04, 104.74]$ \\
Bayes utility oracle & $98.63\,[97.77, 99.31]$ & $118.08\,[113.63, 122.00]$ \\
\bottomrule
\end{tabular}

\end{table}

\subsection{Performance under different action rules and feedback}
Using the same state predictions, the direct policy reached \DirectLate\% late accuracy. The adaptive-minus-direct accuracy difference was \DirectLateDiff\ percentage points. The adaptive state learner approached less often in threat (\AdaptiveUnsafe\% versus \DirectUnsafe\%) and avoided more often in support (\AdaptiveMissed\% versus \DirectMissed\%). Both differences follow the higher approach threshold used by the adaptive state learner. Its cumulative utility difference from the direct policy was \DirectUtilityDiff.

The two value learners fit action payoffs directly. The full-information learner, which derives both action targets from the same observed labels as the logistic learners, reached \ValueLate\% late accuracy. Its late missed-support rate was \ValueMissed\%, compared with \AdaptiveMissed\% for the adaptive state learner. The contextual bandit reached \BanditLate\% late accuracy and \BanditUtility\ cumulative utility, compared with \AdaptiveUtility\ cumulative utility for the adaptive state learner. Adaptive-minus-bandit differences were \BanditLateDiff\ percentage points and \BanditUtilityDiff\ in cumulative utility (Table~\ref{tab:contrasts}).

\subsection{The tested history weights lowered cumulative utility}
The history-weight variant uses the same state predictions as the neutral and fixed-weight variants. Relative to neutral weights, the history variant had a late accuracy difference of \HistoryLateDiff\ percentage points and a cumulative utility difference of \HistoryUtilityDiff. The history variant also earned less cumulative utility than the fixed balanced variant (difference \FixedUtilityDiff).

\begin{figure}[!ht]
\centering\includegraphics[width=\linewidth]{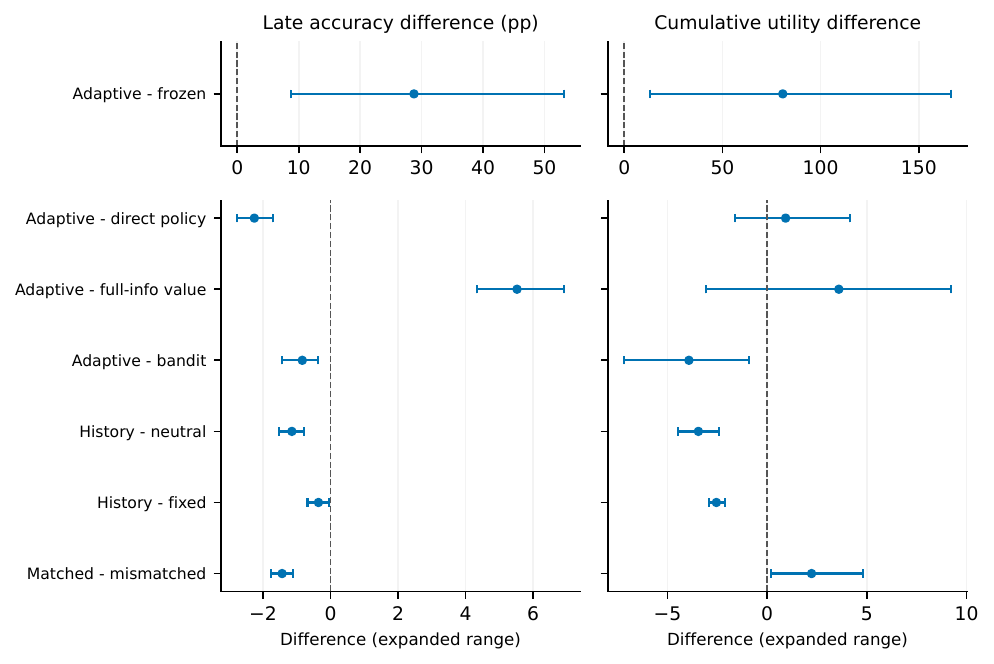}
\caption{Paired mean differences and their 95\% environment-bootstrap intervals. The top row shows the comparison with the frozen learner. Separate horizontal scales in the lower row make the six smaller contrasts visible. Each column uses the units given above it.}
\label{fig:effects}
\end{figure}

Using the actual current weights improved cumulative utility over deliberately mismatched weights by \MatchedUtilityDiff, while lowering late accuracy. The same prediction can thus lead to different outcomes depending on which errors the rule penalizes. Table~\ref{tab:contrasts} and Figure~\ref{fig:effects} show these changes alongside the gain from online updating.

\begin{table}[!ht]
\centering\small
\caption{Paired differences between model conditions with 95\% environment-bootstrap intervals. Each entry subtracts the right-hand condition from the left-hand condition within each of 1,296 units before averaging by environment. Accuracy differences are in percentage points.}
\label{tab:contrasts}
\begin{tabular}{@{}lrr@{}}
\toprule
Left $-$ right & Accuracy difference (pp) & Utility difference \\
\midrule
Adaptive $-$ frozen & $28.76\,[8.73, 53.19]$ & $80.71\,[13.23, 166.42]$ \\
Adaptive $-$ direct policy & $-2.26\,[-2.76, -1.71]$ & $0.92\,[-1.60, 4.15]$ \\
Adaptive $-$ full-information value & $5.52\,[4.33, 6.91]$ & $3.59\,[-3.07, 9.22]$ \\
Adaptive $-$ contextual bandit & $-0.84\,[-1.45, -0.37]$ & $-3.93\,[-7.17, -0.89]$ \\
History $-$ neutral weights & $-1.15\,[-1.52, -0.79]$ & $-3.45\,[-4.48, -2.43]$ \\
History $-$ fixed weights & $-0.36\,[-0.68, -0.06]$ & $-2.55\,[-2.93, -2.09]$ \\
Matched $-$ mismatched weights & $-1.44\,[-1.76, -1.10]$ & $2.22\,[0.20, 4.81]$ \\
\bottomrule
\end{tabular}

\end{table}

\FloatBarrier
\section{Discussion}
The decrease in Brier score shows that continued supervised updates reduced probability error under the changed cue distribution. The task is favorable to a logistic predictor in one respect. At a given episode, the two state-conditional cue distributions have opposite means and equal covariance, so the true log odds are linear in the cues. Transfer changes the coefficients of this relationship and can also change the intercept as the state prior drifts. The six configurations combine changes to cues, priors, and feedback, so their results do not estimate the effect of each perturbation in isolation.

The action results show why state predictions and decision rules need separate evaluation. The direct policy's higher accuracy came from a lower approach threshold and fewer missed support episodes. Its lower point estimate for cumulative utility comes with an interval for the difference that spans zero. Likewise, the history weights changed the decision threshold without changing predictions. This particular rule carried a summary of past exposure into a setting where the current costs varied independently. Its lower cumulative utility addresses that fixed weighting rule. Other ways of using history, such as estimating a changing state prior or learning how costs vary over time, would require their own comparisons.

The bandit comparison concerns feedback content as well as the update rule. The reward was computed from the true state, while three of the six environments flipped some observed labels for supervised learners. The payoff sign gives the bandit an informative, uncorrupted signal about whether its action matched the state. The present design cannot estimate the separate contributions of that signal, the action-specific value updates, and the policy. A comparison designed to isolate feedback access would need to control what can be inferred about the state from each feedback channel.

The exposure manipulation also defines what the task can say about experience. Developmental robotics studies how early experience shapes behavior as an agent interacts with its environment \cite{cangelosi}. Here, early experience varies through the frequency of states during training, and a learner's actions affect only the current payoff. Studying adaptation with the mutual agent--environment coupling modeled by Beer \cite{beer} would require evaluating decisions that alter future observations. The present results provide a comparison for the one-step case with exogenous cues.

\section*{Data and code availability}
The code and synthetic data are available on Zenodo \cite{yin2026archive} at \mbox{\href{https://doi.org/10.5281/zenodo.24329690}{DOI~10.5281/zenodo.24329690}}. The archive contains the configuration, simulator, analysis scripts, and 1,296,000 original transfer outcome records. These records contain the true state, action, realized utility, and the probability estimate where available. A separate export regenerates 69,120 development and 43,200 transfer inputs from the 432 seed pairs. Each input includes the cues and feedback label. Transfer inputs join to outcomes by environment, history, seed index, and episode.

The archive's README gives commands for rerunning the simulation and regenerating the tables and figures. The archived deterministic fields agree with a full rerun at absolute tolerance $10^{-12}$ using Python 3.13.9, NumPy 2.3.5, and pandas 2.3.3. The verification reports trace the tables and plot coordinates to the result files.

\section*{Generative AI disclosure}
OpenAI Codex assisted with drafting and editing the English text, preparing the LaTeX manuscript, and building and running packaging and verification scripts for this version.

\clearpage
\appendix
\section{Cue generation and random streams}\label{app:environment}
Let $\ell_j$ and $\sigma_j$ denote the loading and noise standard deviation of cue $j$ in a given phase. Given the true state $s_t$ of episode $t$, the simulator draws cues independently according to
\begin{equation}
    x_{tj}=(2s_t-1)\ell_j+e_{tj}, \qquad e_{tj}\sim\mathcal{N}(0,\sigma_j^2).
\end{equation}
The support probability is fixed by history during development. In transfer it follows a linear schedule from $p_0$ to $p_1$. The simulator draws states independently across episodes given this schedule.

The generator is seeded with $10{,}000+e$ for a validation environment and $20{,}000+e$ for a test environment, where $e$ is the environment ID. Development loadings are uniform on $[0.65,1.20]$, with the sixth set to 0.05. Development noise standard deviations are uniform on $[0.55,0.90]$. At transfer the generator selects zero, two, or four of the first five cues for sign reversal, then multiplies one loading by 0.20. In even-numbered environments it sets another loading to zero. A zero loading removes the cue's dependence on state while retaining observation noise. Since the generator can zero a selected loading, the selected flip count may exceed the number of sign-reversed cues that remain informative. The sixth cue receives a random sign and a loading magnitude drawn uniformly from $[0.75,1.15]$. Each transfer noise standard deviation is the development value multiplied by a draw from $[0.90,1.30]$.

The generator draws $p_0$ uniformly from $[0.38,0.62]$ and applies an ID-dependent drift of $-0.15$, $0$, or $0.15$ to determine $p_1$, clipped to $[0.20,0.80]$. The state label available for supervised feedback is independently flipped with probability zero or 0.05, depending on the environment. Label flips and an environment-specific delay of zero or two episodes apply in both development and transfer. Table~\ref{tab:environments} lists the test configurations and the resulting late accuracy of the adaptive and frozen learners.

\begin{table}[ht]
\centering\footnotesize
\caption{Held-out environments and mean late action accuracy (\%). The missing index is zero-based. A dash indicates that all cues retain nonzero state loadings. The flip count records selections before weakening or zeroing. Label flip gives the probability of flipping supervised state labels. Delay is measured in episodes.}
\label{tab:environments}
\begin{tabular}{@{}lrrrrrrr@{}}
\toprule
ID & Flips & Missing & $p_0$ to $p_1$ & Label flip & Delay & Adaptive & Frozen \\
\midrule
200 & 4 & 2 & 0.531 to 0.381 & 0.00 & 0 & 91.53 & 39.63 \\
201 & 0 & -- & 0.593 to 0.593 & 0.05 & 0 & 92.43 & 86.39 \\
202 & 2 & 4 & 0.526 to 0.526 & 0.00 & 2 & 95.32 & 89.10 \\
203 & 4 & -- & 0.412 to 0.412 & 0.05 & 2 & 94.07 & 13.70 \\
204 & 0 & 1 & 0.601 to 0.751 & 0.00 & 0 & 93.87 & 85.93 \\
205 & 2 & -- & 0.409 to 0.559 & 0.05 & 0 & 93.12 & 73.06 \\
\bottomrule
\end{tabular}

\end{table}

Within an environment, let $h=0,1,2$ index supportive, ordinary, and adverse histories and let $i$ be the seed index. The environment seed is $20{,}260{,}802+100{,}000e+10{,}000h+i$. The policy seed is the environment seed plus $900{,}000{,}000$. This construction preserves the same observation and exploration streams for all model conditions and payoff profiles within each environment, history, and seed index.

\section{Decision thresholds and update timing}\label{app:updates}
Given an estimate $q$ of the probability of support and the payoff in Table~\ref{tab:payoff}, the expected utility rule prefers approach when
\begin{equation}
    q>\frac{2.45w_s}{2.45w_s+1.35w_a}.
\end{equation}
The matched and frozen conditions use the current utility weights. Neutral weights are $(1,1)$ and fixed balanced weights are $(2,1.5)$. Equal weights still yield an approach threshold above 0.5 because the payoffs are asymmetric. The history rule uses $(1+2(1-\bar{s}_{\mathrm{dev}}),\,1+\bar{s}_{\mathrm{dev}})$, where $\bar{s}_{\mathrm{dev}}$ is the fraction of \emph{true} support states in the 160 development episodes. This gives the rule access to a simulator quantity that could be unavailable when labels are noisy. The resulting weights remain fixed during transfer. Mismatched weights swap the current safety and affiliation weights, including the balanced case $(1.5,2.0)$.

The value learners use the augmented cue vector $\phi$ and one coefficient vector $v_a$ per action. On receipt of a target $r_a$ for action $a$, the update is
\begin{equation}
    v_a\leftarrow v_a+\alpha (r_a-v_a^\top\phi)\phi .
\end{equation}
The full-information learner applies this update to both actions with $r_a=U(\tilde{s},a)$. The bandit applies it only to the selected action, with $r_a=U(s,a)$. Both use balanced payoff weights for development targets and the current profile for transfer targets. The greedy action has the higher predicted value $v_a^\top\phi$.

All learned coefficients start at zero. At the transfer boundary, remaining development feedback is processed before the first transfer decision. During transfer, the learner chooses an action with its current coefficients before processing the feedback that is due in that episode. With delay $d$, the feedback for episode $t$ updates the learner after the decision at $t+d$, so it can first affect the action at $t+d+1$. The frozen learner retains its development coefficients. Transfer feedback still in the queue at the end of transfer has no effect on the reported decisions.

The implementation randomizes numerical ties using shared draws. On exploration episodes the learned conditions use the same draw from the uniform action distribution. The Bayes oracle instead uses the known observation-model parameters and support prior $p_t$ to compute the posterior estimate
\begin{equation}
    q^*=\sigma\!\left(\log\frac{p_t}{1-p_t}
           +2\sum_{j=1}^{6}\frac{\ell_j x_{tj}}{\sigma_j^2}\right).
\end{equation}
It acts on this posterior and the current weights before observing the true state. Sigmoid evaluations clip the logit to $[-30,30]$.

Validation tested mapping rates 0.04, 0.08, and 0.12 and value rates 0.01, 0.03, and 0.05. If two candidates tied on cumulative utility and late accuracy, the lower rate was preferred. The archived validation grid gives each family's candidate scores.

\section{Supplementary outcomes}\label{app:metrics}
Let $c_t$ equal 1 when $a_t=s_t$ and 0 otherwise. The normalized accuracy area is the trapezoidal sum of $c_t$ over all 100 transfer episodes divided by 99. The first threshold crossing is the earliest ending episode of a complete ten-episode window with mean $c_t$ at least 0.8. It is coded as 101 if the threshold is never reached. This metric describes the first successful window, including the possibility of a brief crossing in a run with lower accuracy later.

Table~\ref{tab:errors} reports late conditional action errors. All archived late windows contain at least one episode of each state. Table~\ref{tab:dynamics} reports early accuracy, accuracy area, and first threshold crossing. Table~\ref{tab:stateutility} reports mean utility within each true state and the Brier score. Both use all 100 transfer episodes. The six online logistic conditions produce identical Brier point estimates because they share predictions. Their intervals vary slightly because the analysis draws a separate bootstrap resample sequence for each model and metric.

\begin{table}[tb]
\centering\small
\caption{Late conditional action errors (\%) with 95\% environment-bootstrap intervals. Within each run, the denominators are the numbers of threat and support episodes, respectively, in episodes 81--100.}
\label{tab:errors}
\begin{tabular}{@{}lrr@{}}
\toprule
Model condition & $P(a=1\mid s=0)$ (\%) & $P(a=0\mid s=1)$ (\%) \\
\midrule
Adaptive state & $3.09\,[2.41, 3.84]$ & $10.01\,[7.80, 12.19]$ \\
Frozen state & $25.06\,[4.71, 50.35]$ & $44.95\,[21.92, 70.77]$ \\
Direct policy & $4.12\,[3.24, 5.03]$ & $4.62\,[3.77, 5.59]$ \\
Full-information value & $2.77\,[2.34, 3.34]$ & $20.95\,[17.79, 23.94]$ \\
Contextual bandit & $3.56\,[2.74, 4.73]$ & $8.27\,[6.25, 10.70]$ \\
Neutral weights & $3.24\,[2.64, 3.88]$ & $7.81\,[6.10, 9.84]$ \\
Fixed balanced weights & $2.96\,[2.23, 3.63]$ & $9.78\,[7.52, 12.32]$ \\
History weights & $3.10\,[2.48, 3.78]$ & $10.29\,[7.85, 12.74]$ \\
Mismatched weights & $3.83\,[3.08, 4.59]$ & $6.53\,[5.15, 8.02]$ \\
Bayes utility oracle & $0.84\,[0.38, 1.34]$ & $2.09\,[0.95, 3.96]$ \\
\bottomrule
\end{tabular}

\end{table}

\begin{table}[tb]
\centering\scriptsize\setlength{\tabcolsep}{4pt}
\caption{Early action accuracy, normalized accuracy area, and first ten-episode threshold crossing. Entries give means and 95\% environment-bootstrap intervals. A run that never crosses the threshold contributes 101 to the crossing score.}
\label{tab:dynamics}
\begin{tabular}{@{}lrrr@{}}
\toprule
Model condition & Early accuracy (\%) & Normalized AUC & First threshold crossing \\
\midrule
Adaptive state & $78.01\,[67.37, 88.29]$ & $0.895\,[0.868, 0.920]$ & $13.98\,[10.71, 17.98]$ \\
Frozen state & $64.27\,[39.18, 85.62]$ & $0.648\,[0.402, 0.856]$ & $39.44\,[11.55, 70.59]$ \\
Direct policy & $80.38\,[68.66, 91.84]$ & $0.918\,[0.887, 0.944]$ & $13.19\,[10.27, 16.59]$ \\
Full-information value & $80.50\,[75.64, 84.71]$ & $0.865\,[0.850, 0.882]$ & $12.26\,[11.26, 13.30]$ \\
Contextual bandit & $83.93\,[77.72, 89.82]$ & $0.918\,[0.898, 0.935]$ & $11.96\,[10.58, 13.62]$ \\
Neutral weights & $79.49\,[68.42, 90.19]$ & $0.908\,[0.876, 0.933]$ & $13.48\,[10.26, 17.10]$ \\
Fixed balanced weights & $78.38\,[67.81, 88.68]$ & $0.898\,[0.867, 0.924]$ & $13.76\,[10.55, 17.81]$ \\
History weights & $77.09\,[66.96, 87.18]$ & $0.889\,[0.859, 0.913]$ & $14.23\,[11.00, 17.70]$ \\
Mismatched weights & $79.68\,[68.48, 90.47]$ & $0.911\,[0.879, 0.937]$ & $13.41\,[10.35, 17.43]$ \\
Bayes utility oracle & $98.61\,[97.63, 99.48]$ & $0.987\,[0.979, 0.994]$ & $10.01\,[10.00, 10.03]$ \\
\bottomrule
\end{tabular}

\end{table}

\begin{table}[tb]
\centering\scriptsize\setlength{\tabcolsep}{4pt}
\caption{Full-transfer state-conditional mean utilities and Brier scores, with 95\% environment-bootstrap intervals. A dash marks an undefined Brier score for a learner that outputs values without a probability estimate.}
\label{tab:stateutility}
\begin{tabular}{@{}lrrr@{}}
\toprule
Model condition & Utility given threat & Utility given support & Brier score \\
\midrule
Adaptive state & $0.66\,[0.58, 0.73]$ & $1.18\,[1.08, 1.28]$ & $0.0500\,[0.0296, 0.0741]$ \\
Frozen state & $-0.30\,[-1.48, 0.67]$ & $0.61\,[0.07, 1.09]$ & $0.2473\,[0.0618, 0.4714]$ \\
Direct policy & $0.50\,[0.36, 0.63]$ & $1.32\,[1.25, 1.39]$ & $0.0500\,[0.0297, 0.0739]$ \\
Full-information value & $0.74\,[0.70, 0.77]$ & $1.05\,[0.98, 1.13]$ & -- \\
Contextual bandit & $0.64\,[0.60, 0.68]$ & $1.27\,[1.20, 1.34]$ & -- \\
Neutral weights & $0.62\,[0.53, 0.71]$ & $1.23\,[1.13, 1.32]$ & $0.0500\,[0.0297, 0.0747]$ \\
Fixed balanced weights & $0.66\,[0.58, 0.74]$ & $1.18\,[1.06, 1.28]$ & $0.0500\,[0.0294, 0.0735]$ \\
History weights & $0.63\,[0.55, 0.71]$ & $1.15\,[1.04, 1.25]$ & $0.0500\,[0.0294, 0.0742]$ \\
Mismatched weights & $0.54\,[0.41, 0.64]$ & $1.26\,[1.16, 1.34]$ & $0.0500\,[0.0297, 0.0734]$ \\
Bayes utility oracle & $0.86\,[0.84, 0.89]$ & $1.46\,[1.43, 1.48]$ & $0.0085\,[0.0040, 0.0134]$ \\
\bottomrule
\end{tabular}

\end{table}

\end{document}